\documentclass{SPAICE}

\def\authorEmail{miguel.leiva.velez@alumnos.upm.es}
\def\AuthorShort{M. Leiva-Vélez et al.}

\author[1,2]{Miguel Leiva-Vélez\thanks{Corresponding author. E-Mail: \authorEmail}}
\author[2]{Nicolas Gaston-Rozado}
\author[3]{Andrea Scorsoglio}
\author[4]{Hodei Urrutxua}
\author[5]{Víctor Rodríguez-Fernández}
\author[3]{Roberto Furfaro}
\affil[1]{ETSI Aeronáutica y del Espacio, Universidad Politécnica de Madrid, Madrid, Spain}
\affil[2]{Indra Sistemas, Madrid, Spain}
\affil[3]{Department of Systems and Industrial Engineering, University of Arizona, Tucson, USA}
\affil[4]{Escuela de Ingeniería de Fuenlabrada, Universidad Rey Juan Carlos, Fuenlabrada, Spain}
\affil[5]{Department of Computer Systems Engineering, Universidad Politécnica de Madrid, Madrid, Spain}

\title{Adversarial Reinforcement Learning for Robust Planetary Landing}

\begin{document}

\maketitle

\begin{abstract}

Deep reinforcement learning (DRL) policies for planetary landing and other aerospace guidance, navigation, and control tasks are commonly trained under randomized disturbances to improve robustness. However, typical stochastic perturbation sampling assumes independent variations, whereas real mission disturbances often exhibit spatial and temporal correlation and cannot be reproduced through independent random draws. This work investigates adversarial reinforcement learning to discover challenging off-nominal conditions. A secondary agent generates physically constrained perturbations that degrade the performance of a landing controller, forming an asymmetric sequential zero-sum game. Training proceeds in three stages: baseline controller learning, robustness enhancement with stochastic perturbations, and an iterative adversarial phase in which the controller and disturbance generator are alternately optimized. Preliminary experiments in a three-degree-of-freedom landing environment show that adversarial disturbances reveal vulnerabilities missed by conventional randomization and provide early indications that, with a more extensive training campaign, the approach can yield improved robustness and stronger generalization under previously unseen perturbation regimes.

\end{abstract}

\section{Introduction}

Autonomous planetary landing is a safety-critical guidance, navigation, and control (GNC) problem characterized by nonlinear dynamics, limited control authority, and significant environmental and parametric uncertainty. During descent, the vehicle must satisfy strict terminal constraints while maintaining safety and fuel efficiency under adverse conditions, making robust performance essential.

Deep reinforcement learning (DRL) has emerged as a promising framework for autonomous landing, enabling direct learning of closed-loop control policies in simulated environments and handling nonlinear dynamics and long-horizon objectives. To improve robustness, most DRL approaches introduce stochastic perturbations or domain randomization during training.

While effective for improving average-case performance, stochastic sampling does not necessarily capture mission-relevant disturbance patterns, which often evolve sequentially, interact with vehicle dynamics and flight phase, and exhibit spatial or temporal correlation. Because such structured scenarios occupy only a small subset of the disturbance space, they are unlikely to be discovered through independent random sampling, leaving potential failure modes unexplored.

This work proposes adversarial reinforcement learning to explore challenging disturbance scenarios in planetary landing. A secondary agent learns to generate temporally correlated perturbation sequences that degrade controller performance. The interaction between the landing controller and disturbance generator is formulated as an asymmetric sequential zero-sum game, allowing the adversary to search a high-dimensional disturbance space for difficult yet physically plausible scenarios.

The approach is evaluated in a three-degree-of-freedom lunar landing environment. Early results indicate that adversarially generated disturbances expose vulnerabilities not captured by conventional stochastic randomization and potentially lead to policies that remain effective under previously unseen and more severe perturbation regimes.

\section{Related Work}

Planetary landing GNC remains dominated in practice by classical guidance and model-based optimal control due to their interpretability and reliability. ZEM/ZEV-type guidance laws provide compact solutions for nominal scenarios \cite{furfaro2020adaptive}, while convex optimization-based powered descent enables real-time trajectory generation with explicit constraint handling and strong theoretical guarantees \cite{blackmore2010minimum,accikmecse2013lossless}.

Recent academic work has explored DRL as a complementary approach for handling nonlinear dynamics and uncertain environments. Prior studies have demonstrated DRL-based powered descent control, algorithm comparisons, adaptive recurrent policies, hybrid classical-learning guidance architectures, transfer across planetary environments, and landing site selection under divert constraints \cite{gaudet2018deep,xu2021deep,gaudet2019adaptive,furfaro2020adaptive,ciabatti2021autonomous,iiyama2021deep}. While promising, these approaches remain sensitive to reward design and distribution shift.

As already stated, most robustness-oriented DRL methods rely on stochastic perturbation sampling or domain randomization \cite{gaudet2019adaptive,xu2021deep,ciabatti2021autonomous}. Although effective for improving average-case resilience, such approaches may under-sample rare but mission-relevant disturbance sequences. Adversarial reinforcement learning addresses this limitation by modeling disturbance generation as a sequential decision process that actively searches for performance-degrading scenarios \cite{pinto2017robust}. This work investigates the application of this idea to DRL-based planetary landing.

\section{Problem Formulation}

We consider autonomous planetary landing in a three-degree-of-freedom setting, where the lander is modeled as a point mass in a target-centered inertial frame with state
\begin{equation}
\mathbf{x}(t)=
\begin{bmatrix}
\mathbf{r}(t)\\
\mathbf{v}(t)\\
m(t)
\end{bmatrix},
\label{eq:state}
\end{equation}
where $\mathbf{r}\in\mathbb{R}^3$ is position, $\mathbf{v}\in\mathbb{R}^3$ is velocity, and $m\in\mathbb{R}^{+}$ is mass. The translational dynamics are
\begin{align}
\dot{\mathbf{r}} &= \mathbf{v}\,, \\
\dot{\mathbf{v}} &= \mathbf{g}(\mathbf{r}) + \frac{1}{m}\mathbf{T}_{\mathrm{eff}} + \mathbf{d}_{\mathrm{ext}}(\mathbf{x},t)\,, \\
\dot{m} &= -\frac{\|\mathbf{T}\|}{I_{\mathrm{sp}} g_0}\,,
\label{eq:dynamics}
\end{align}
where $\mathbf{T}$ is the commanded thrust, $\mathbf{g}(\mathbf{r})$ is the local gravity field, and $I_{\mathrm{sp}}$ is the specific impulse. Internal actuation imperfections are captured through
\begin{equation}
\mathbf{T}_{\mathrm{eff}}=\mathbf{T}+\mathbf{d}_{\mathrm{int}}(\mathbf{x},\mathbf{T},t)\,,
\label{eq:effective_thrust}
\end{equation}
where $\mathbf{d}_{\mathrm{int}}$ models effects such as thrust bias, misalignment, or degradation, and $\mathbf{d}_{\mathrm{ext}}$ represents external disturbances. The objective is to learn a policy that achieves soft landing conditions,
\begin{equation}
\mathbf{r}(t_f)\approx \mathbf{0}\,, \qquad \mathbf{v}(t_f)\approx \mathbf{0}\,,
\label{eq:terminal}
\end{equation}
while respecting mission and control constraints throughout descent.

The controller operates under uncertainty and partial observability. At each time step, it receives a noisy observation
\begin{equation}
\mathbf{o}_t=h(\mathbf{x}_t)+\boldsymbol{\eta}_t\,,
\label{eq:observation}
\end{equation}
where $h(\cdot)$ is the observation model and $\boldsymbol{\eta}_t$ is sensor noise. The landing task is therefore modeled as a partially observable Markov decision process,
\begin{equation}
(\mathcal{S},\mathcal{A},\mathcal{O},P,R,\gamma)\,,
\label{eq:pomdp}
\end{equation}
with continuous thrust actions and reward function $R$ designed to promote safe, accurate, and fuel-efficient descent. The controller seeks a policy maximizing the expected return
\begin{equation}
\mathbb{E}\left[\sum_{t=0}^{t_f}\gamma^t R(\mathbf{x}_t,\mathbf{T}_t)\right].
\label{eq:return}
\end{equation}

Perturbation generation is modeled as a sequential decision process in which, alongside the landing controller, a second agent acts as an adversary and selects physically admissible disturbances over time. Their interaction is formulated as an asymmetric zero-sum game: the protagonist seeks successful landing, while the adversary aims to maximally degrade performance through bounded perturbation sequences. Denoting the returns of the controller and adversary by $R_A$ and $R_B$, respectively,
\begin{equation}
R_A + R_B = 0\,.
\label{eq:zerosum}
\end{equation}

\section{Methodology}

We study robust lunar landing in a simplified three-degree-of-freedom (3-DOF) environment, adopted here as an intermediate step toward subsequent extension to full six-degree-of-freedom dynamics. Training is organized as a staged curriculum to promote stable policy acquisition under progressively more challenging conditions. A baseline controller is first optimized in the nominal environment with randomized initial conditions. This policy is then fine-tuned under stochastic perturbations of increasing intensity. In the final stage, an adversarial agent is introduced and trained iteratively against the robustified controller to generate disturbance sequences that maximally deteriorate landing performance.

The lander policy produces a continuous four-dimensional control action,
\begin{equation}
\mathbf{a}_L \in [-1,1]^4,
\end{equation}
whose first three components specify thrust direction, subsequently normalized to unit norm, and whose fourth component sets thrust magnitude. During adversarial training, a second agent acts in the disturbance space by controlling three perturbation channels: the magnitude $F_{\mathrm{ext}}$ and direction $\hat{\mathbf{f}}_{\mathrm{ext}}$ of an external force, the thrust misalignment angle $\theta_{\mathrm{mis}}$, and the specific-impulse degradation $\ell_{\mathrm{isp}}$. These perturbations are constrained to the same bounds used during stochastic robustness training. Rather than selecting absolute disturbance values at each time step, the adversary applies bounded increments, allowing perturbations to evolve smoothly over time and thereby preserving transient dynamics while avoiding unrealistic step changes.

Guidance is induced through a position-dependent reference velocity field $\mathbf{v}_{\mathrm{ref}}(\mathbf{r})$, which steers the vehicle toward the landing site and transitions to a near-vertical descent in the terminal phase. At each step, reward is based on progress projected along the local streamline direction,
\begin{equation}
\hat{\mathbf{v}}_{\mathrm{ref}}=
\frac{\mathbf{v}_{\mathrm{ref}}(\mathbf{r}_{t-1})}{\|\mathbf{v}_{\mathrm{ref}}(\mathbf{r}_{t-1})\|},
\end{equation}
with instantaneous progress
\begin{equation}
p_t = (\mathbf{r}_t-\mathbf{r}_{t-1})\cdot \hat{\mathbf{v}}_{\mathrm{ref}},
\end{equation}
and admissible progress
\begin{equation}
p_t^{\star}=\max\left(\|\mathbf{v}_{\mathrm{ref}}(\mathbf{r}_t)\|,v_{\mathrm{floor}}\right)\Delta t.
\end{equation}
The progress term is defined as
\begin{equation}
r_t^{\mathrm{prog}}=
\begin{cases}
p_t, & p_t \leq p_t^{\star},\\
2p_t^{\star}-p_t, & p_t > p_t^{\star},
\end{cases}
\end{equation}
so that motion beyond the admissible local rate is penalized. The resulting lander reward is
\begin{equation}
r_t^{L}=
\frac{r_t^{\mathrm{prog}}}{p_t^{\star}+\varepsilon}
-\lambda_T \frac{\|\mathbf{T}_t\|}{T_{\max}}.
\end{equation}
The adversary, by contrast, is trained using only a sparse terminal objective proportional to final landing error,
\begin{equation}
R_{\mathrm{adv}}=\|\mathbf{r}(t_f)-\mathbf{r}_{\mathrm{pad}}\|,
\end{equation}
which concentrates learning on degradation of terminal performance without inheriting the protagonist reward shaping.

The observation vector supplied to the lander comprises vehicle and guidance variables $(\mathbf{r}-\mathbf{r}_{\mathrm{pad}},\mathbf{v}-\mathbf{v}_{\mathrm{ref}},\|\mathbf{T}_{t-1}\|,\hat{\mathbf{T}}_{t-1})$, internal perturbation states $(\theta_{\mathrm{mis}},\ell_{\mathrm{isp}})$, and external perturbation variables $(\mathbf{a}_{\mathrm{pert}},\|\mathbf{a}_{\mathrm{pert}}\|)$. All channels are normalized prior to network input. The adversary receives the same information, but encoded through a recurrent state to capture temporal correlations across the trajectory. 

Performance is assessed under two perturbation regimes: in-distribution (ID) evaluation, in which disturbance magnitudes remain within the training range, and out-of-distribution (OOD) evaluation, in which larger unseen perturbations are applied to test generalization. Robustness is analyzed by systematically sweeping the principal disturbance axes (external force magnitude, thrust misalignment, and specific-impulse efficiency loss) both independently and in combination, thereby probing the controller’s sensitivity to individual perturbations as well as to compounded degradation effects. For each configuration, disturbance directions are randomized at the beginning of each episode to avoid directional bias. Evaluation is performed over one hundred Monte Carlo rollouts per scenario to capture statistical variability in the resulting trajectories and landing outcomes, enabling consistent estimates of performance under progressively increasing perturbation levels and across multi-axis disturbance conditions.

Algorithmically, Proximal Policy Optimization (PPO) \cite{schulman2017proximal} is the selected DRL paradigm both for the main policy and the adversarial one.

\section{Results}

\begin{figure}[h!]
\centering
\includegraphics[width=0.95\columnwidth]{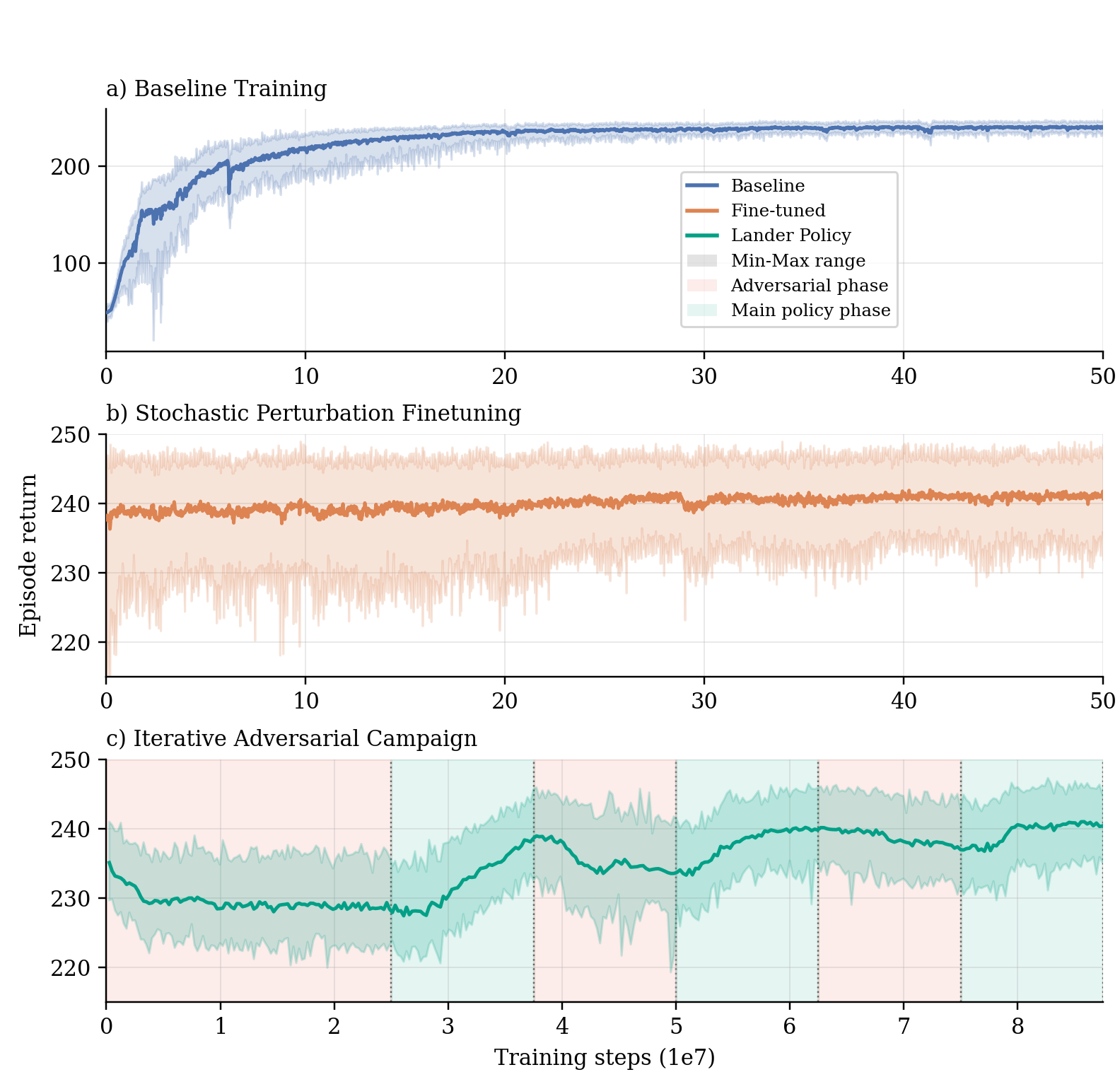}
\caption{Training curves across the curriculum}
\label{fig:training_curves}
\end{figure}

We first analyze learning behavior in the nominal environment. As shown in \cref{fig:training_curves}, the baseline policy trained with PPO exhibits fast convergence and stable learning despite randomized initial conditions. This indicates that the reward formulation provides a consistent training signal and enables the policy to generalize across the nominal state distribution.

\begin{figure}[h!]
\centering
\includegraphics[width=0.75\columnwidth]{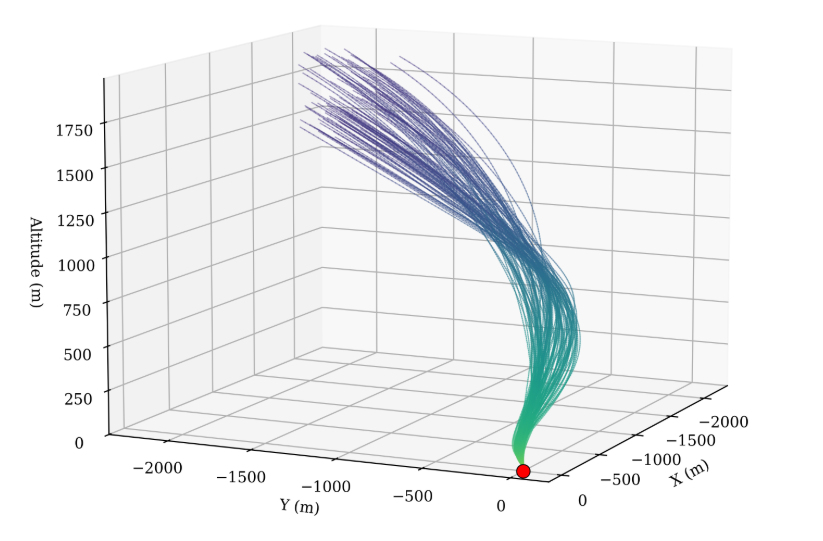}
\caption{Representative 3D landing trajectories.}
\label{fig:trajectories}
\end{figure}

Representative evaluation trajectories are shown in \cref{fig:trajectories}. The policy learns a stable descent geometry and consistently reaches the landing region under nominal conditions.

During stochastic fine-tuning, the agent rapidly compensates for the baseline-policy performance degradation introduced by perturbations, which explains the relatively small variation visible in subplot b) of \cref{fig:training_curves}. Nevertheless, increasing disturbance levels still reduce landing precision, particularly in the terminal phase.

Introducing the adversarial agent exposes disturbance sequences that significantly degrade the stochastically finetuned frozen policy, even within the same perturbation bounds used during stochastic training. The alternating dynamics visible subplot c) in \cref{fig:training_curves} show the expected exploit-adapt pattern, where the adversary first degrades the policy and the protagonist subsequently learns to counteract the discovered perturbation strategy.

\begin{figure}[h!]
\centering
\includegraphics[width=1.0\columnwidth]{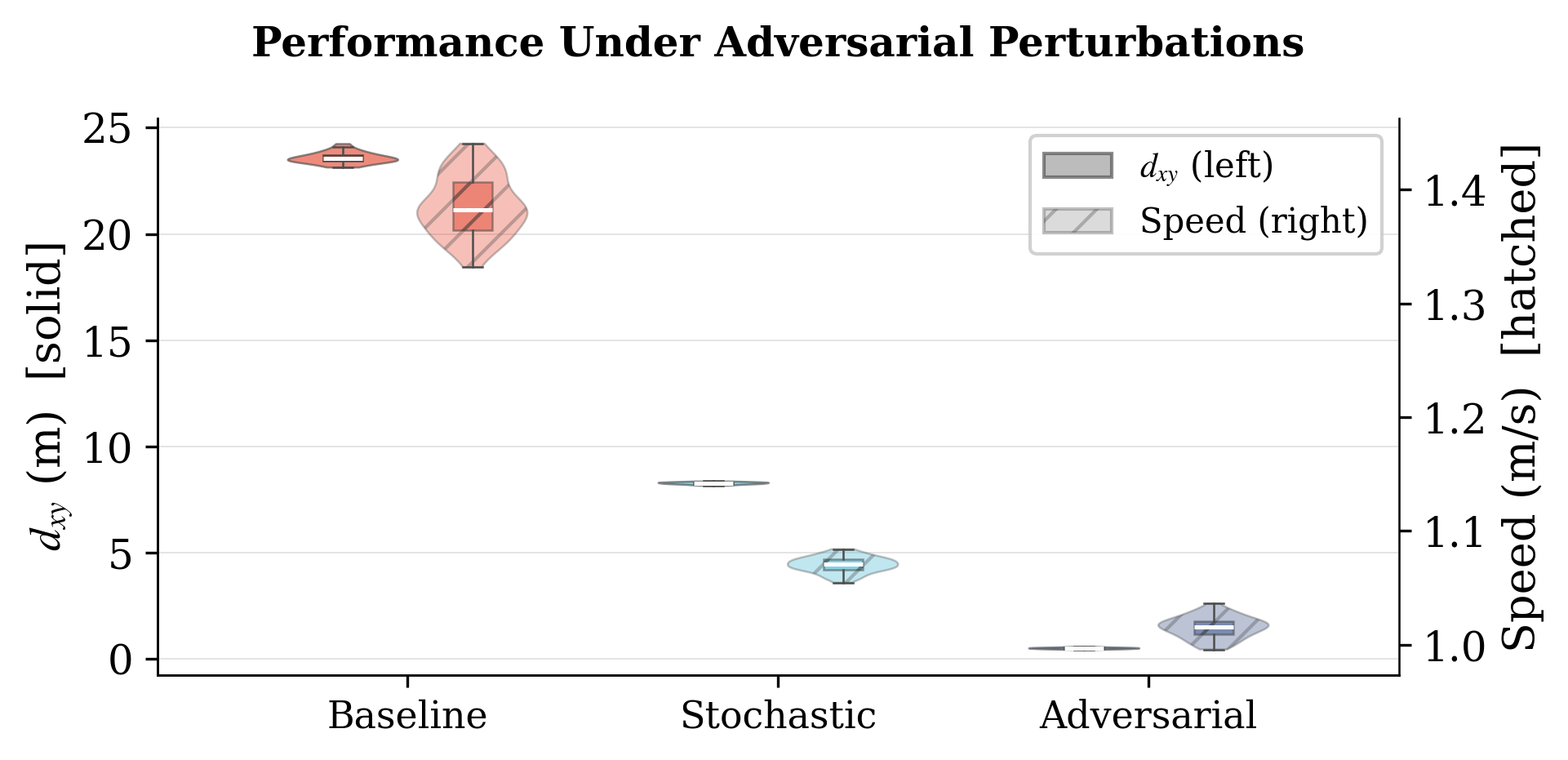}
\caption{Violin plot comparing landing performance of baseline, fine-tuned, and adversarially trained policies against adversarial disturbances.}
\label{fig:vsadversarial}
\end{figure}

This effect is illustrated in \cref{fig:vsadversarial}, where the stochastically fine-tuned policy shows degraded landing precision when evaluated against the adversary, if compared to the adversarially-hardened one.

\begin{figure}[h!]
\centering
\includegraphics[width=0.95\columnwidth]{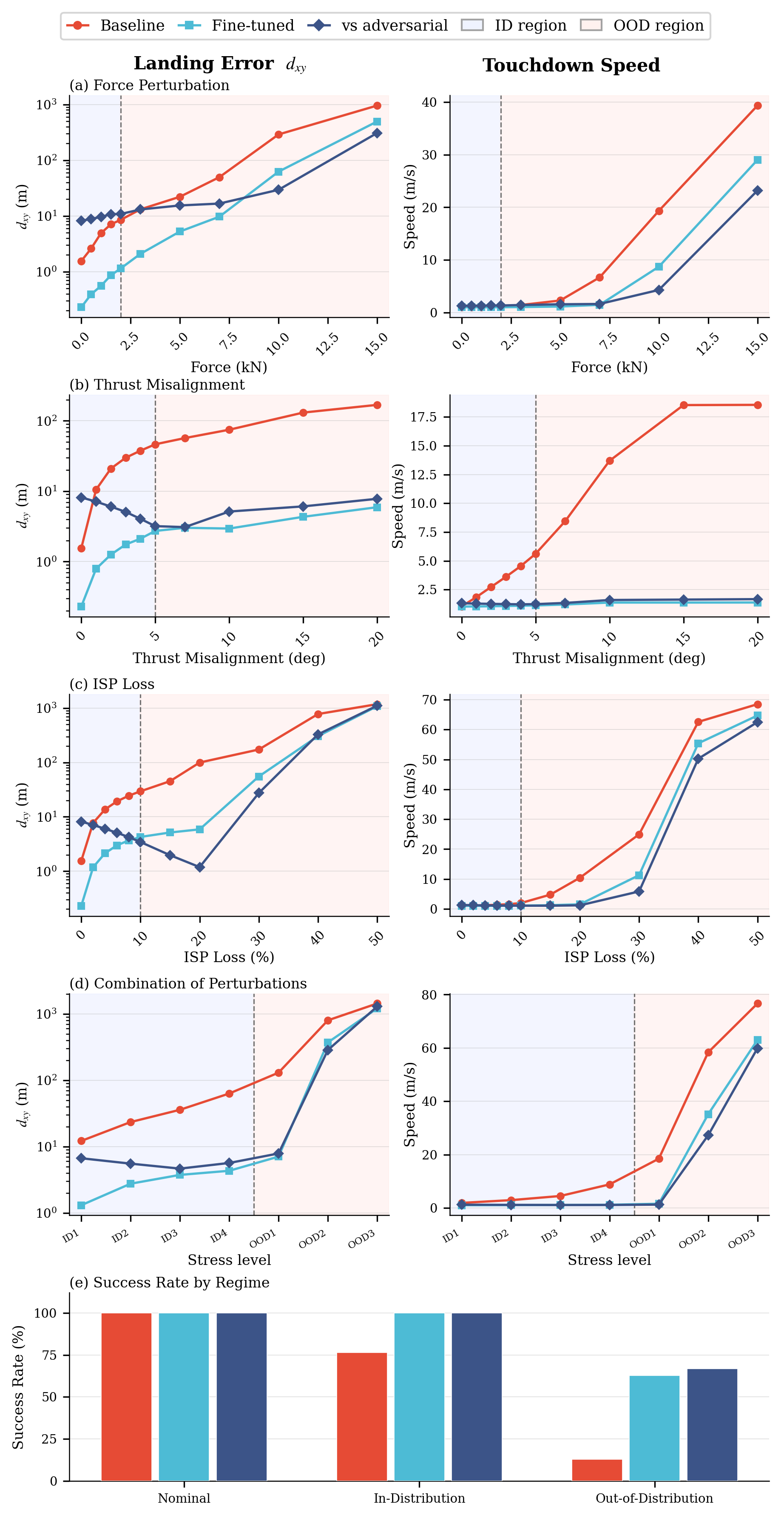}
\caption{Monte Carlo evaluation of baseline, stochastic fine-tuned, and adversarially trained policies under increasing perturbation levels applied individually and jointly.}
\label{fig:combined}
\end{figure}

Then, robustness and generalizability across perturbation levels is evaluated through a Monte Carlo campaign where each policy is tested under progressively increasing disturbances applied individually and jointly. The results are summarized in \cref{fig:combined}. The baseline controller degrades immediately once perturbations are introduced. The stochastically fine-tuned policy shows a clear improvement across most perturbation conditions. The adversarially trained controller demonstrates stronger robustness at higher disturbance levels, although some degradation appears in lower-perturbation regions.

\section{Discussion}

These results highlight three main points. First, the nominal PPO policy learns a stable and generalizable descent strategy under randomized initial conditions, indicating that the proposed reward formulation is sufficient to shape the landing behavior without requiring additional guidance heuristics. This is an important starting point, since the subsequent robustness analyses depend on the existence of a well-behaved baseline controller.

Second, stochastic perturbation training clearly improves robustness relative to the nominal policy, but it does not fully characterize the vulnerability space of the controller. Although the fine-tuned agent performs well under randomly sampled disturbances, the adversarial experiments show that temporally correlated perturbation sequences within the same admissible bounds can still induce substantial degradation. This suggests that robustness assessed only through random sampling may overestimate the true resilience of the policy, particularly in safety-critical phases such as terminal descent.

Third, the initial adversarial campaign demonstrates both the promise and the limitation of adversarial reinforcement learning in this setting. On one hand, adversarial training exposes failure modes that stochastic training alone does not reveal, and improves performance under stronger perturbations. On the other hand, the current results also indicate partial specialization to the adversarial regime, with some loss of precision in milder conditions. This points to the need for a more balanced training procedure in which adversarial, stochastic, and nominal scenarios are continuously interleaved to avoid overfitting to a narrow subset of disturbance realizations.

Overall, this study should be interpreted as a preliminary but encouraging step toward robust RL-based planetary landing. The results support the use of adversarial agents as active disturbance discoverers, but also indicate that robust policy synthesis will likely require broader training schemes, such as mixed-distribution curricula or league-based training inspired by large-scale multi-agent reinforcement learning systems such as AlphaStar~\cite{vinyals2019grandmaster}, to achieve uniform performance across the full perturbation envelope.

\newpage

\printbibliography

@article{blackmore2010minimum,
  title={Minimum-landing-error powered-descent guidance for Mars landing using convex optimization},
  author={Blackmore, Lars and A{\c{c}}ikme{\c{s}}e, Beh{\c{c}}et and Scharf, Daniel P},
  journal={Journal of guidance, control, and dynamics},
  volume={33},
  number={4},
  pages={1161--1171},
  year={2010}
}

@article{accikmecse2013lossless,
  title={Lossless convexification of nonconvex control bound and pointing constraints of the soft landing optimal control problem},
  author={A{\c{c}}{\i}kme{\c{s}}e, Beh{\c{c}}et and Carson, John M and Blackmore, Lars},
  journal={IEEE transactions on control systems technology},
  volume={21},
  number={6},
  pages={2104--2113},
  year={2013},
  publisher={IEEE}
}

@article{gaudet2018deep,
  title={Deep reinforcement learning for six degree-of-freedom planetary powered descent and landing},
  author={Gaudet, Brian and Linares, Richard and Furfaro, Roberto},
  journal={arXiv preprint arXiv:1810.08719},
  year={2018}
}

@article{gaudet2019adaptive,
  title={Adaptive guidance with reinforcement meta-learning},
  author={Gaudet, Brian and Linares, Richard},
  journal={arXiv preprint arXiv:1901.04473},
  year={2019}
}

@article{furfaro2020adaptive,
  title={Adaptive generalized ZEM-ZEV feedback guidance for planetary landing via a deep reinforcement learning approach},
  author={Furfaro, Roberto and Scorsoglio, Andrea and Linares, Richard and Massari, Mauro},
  journal={Acta Astronautica},
  volume={171},
  pages={156--171},
  year={2020},
  publisher={Elsevier}
}

@inproceedings{ciabatti2021autonomous,
  title={Autonomous planetary landing via deep reinforcement learning and transfer learning},
  author={Ciabatti, Giulia and Daftry, Shreyansh and Capobianco, Roberto},
  booktitle={Proceedings of the IEEE/CVF Conference on Computer Vision and Pattern Recognition},
  pages={2031--2038},
  year={2021}
}

@article{iiyama2021deep,
  title={Deep reinforcement learning for safe landing site selection with concurrent consideration of divert maneuvers},
  author={Iiyama, Keidai and Tomita, Kento and Jagatia, Bhavi A and Nakagawa, Tatsuwaki and Ho, Koki},
  journal={arXiv preprint arXiv:2102.12432},
  year={2021}
}

@article{xu2021deep,
  title={Deep reinforcement learning-based accurate control of planetary soft landing},
  author={Xu, Xibao and Chen, Yushen and Bai, Chengchao},
  journal={Sensors},
  volume={21},
  number={23},
  pages={8161},
  year={2021},
  publisher={MDPI}
}

@inproceedings{pinto2017robust,
  title={Robust adversarial reinforcement learning},
  author={Pinto, Lerrel and Davidson, James and Sukthankar, Rahul and Gupta, Abhinav},
  booktitle={International conference on machine learning},
  pages={2817--2826},
  year={2017},
  organization={PMLR}
}

@article{schulman2017proximal,
  title={Proximal policy optimization algorithms},
  author={Schulman, John and Wolski, Filip and Dhariwal, Prafulla and Radford, Alec and Klimov, Oleg},
  journal={arXiv preprint arXiv:1707.06347},
  year={2017}
}

@article{vinyals2019grandmaster,
  title={Grandmaster level in StarCraft II using multi-agent reinforcement learning},
  author={Vinyals, Oriol and Babuschkin, Igor and Czarnecki, Wojciech M and Mathieu, Micha{\"e}l and Dudzik, Andrew and Chung, Junyoung and Choi, David H and Powell, Richard and Ewalds, Timo and Georgiev, Petko and others},
  journal={nature},
  volume={575},
  number={7782},
  pages={350--354},
  year={2019},
  publisher={Nature Publishing Group UK London}
}
\addcontentsline{toc}{section}{References}

\end{document}